\documentclass[runningheads]{llncs}
\usepackage[T1]{fontenc}
\usepackage{graphicx,verbatim}
\usepackage{booktabs,multirow, amsmath, amsfonts, float, hyperref, lineno}

\usepackage{orcidlink}

\begin{document}
%
\title{PPS-Ctrl: Controllable Sim-to-Real Translation for Colonoscopy Depth Estimation}

\author{Xinqi~Xiong\inst{1} \and
Andrea~Dunn~Beltran\inst{1} \and
Jun~Myeong~Choi\inst{1} \and
Marc~Niethammer\inst{2} \and
Roni~Sengupta\inst{1}}
\authorrunning{X. Xiong et al.}
%
\institute{University of North Carolina at Chapel Hill, Chapel Hill, USA
\and
University of California, San Diego, USA}


\maketitle              
\begin{abstract}


Accurate depth estimation enhances endoscopy navigation and diagnostics, but obtaining ground-truth depth in clinical settings is challenging. Synthetic datasets are often used for training, yet the domain gap limits generalization to real data. We propose a novel image-to-image translation framework that preserves structure while generating realistic textures from clinical data. Our key innovation integrates Stable Diffusion with ControlNet, conditioned on a latent representation extracted from a Per-Pixel Shading (PPS) map. PPS captures surface lighting effects, providing a stronger structural constraint than depth maps. Experiments show our approach produces more realistic translations and improves depth estimation over GAN-based MI-CycleGAN. Our code is publicly accessible at \href{https://github.com/anaxqx/PPS-Ctrl}{https://github.com/anaxqx/PPS-Ctrl}.


\end{abstract}

\section{Introduction}

Colorectal cancer is among the most lethal malignancies globally~\cite{araghi2019global}, and colono-scopy remains the gold standard to detect intestinal lesions, such as polyps and colon cancer~\cite{rex2015quality}. 
Accurate depth estimation in colonoscopy enhances spatial understanding and procedural guidance, playing a crucial role in developing assistive navigation and diagnostic tools. 
Improving estimation accuracy aids in guiding exploration to unsurveyed regions~\cite{wu2024toder}, detecting blind spots~\cite{kim2024density} and polyps~\cite{wan2021polyp}, and robotic navigation systems~\cite{wu2024toder,beltran2024nfl}.


As most endoscopes do not support high-quality 3D sensing, clinical colono-scopy datasets often lack ground-truth depth annotations. Consequently, most deep learning-based depth estimation models rely on synthetic datasets where depth annotations can be easily generated~\cite{cheng2021depth,jeong2021depth,ppsnet}. However, synthetic datasets often lack fine-grained textures, realistic lighting conditions, and complex anatomical structures, such as mucus-covered surfaces and polyp variations, present in real clinical data, making it difficult for models trained on synthetic data to generalize to real clinical data. 


Sim-to-real image translation has been explored to mitigate this gap, primarily through GAN-based methods \cite{wang2024structure,jeong2021depth}. However, GANs struggle with mode collapse, structural inconsistencies, and unstable training. Diffusion models offer a more stable alternative\cite{rombach2022high}, but they require stronger constraints to preserve structure in complex environments like endoscopy.

We propose a novel sim-to-real image translation framework that leverages a diffusion model instead of GAN and introduces the Per-Pixel Shading (PPS) map \cite{ppsnet} as a structural constraint, instead of a depth map, for guiding the Stable Diffusion model with a ControlNet. We further introduce an encoder-decoder architecture that extracts more meaningful structural information from PPS map as input to the ControlNet. Per-Pixel Shading map calculates the effect of lighting at each point on the surface defined by the depth map and accounts for the fact that points closer to the endoscope and facing the endoscope should receive a larger intensity of light compared to points further away. Explicit conditioning with the PPS map instead of depth leads to better alignment between the translated image and the structure of the source image while improving realism. 



To summarize, the contributions of our work are as follows:
\begin{itemize}
\item We show that Per-Pixel Shading (PPS) map is a better structural constraint for sim-to-real translation than depth map. 
\item We propose a novel structure-preserving sim-to-real image translation framework using Stable Diffusion and ControlNet, incorporating an encoder-decoder architecture trained to extract meaningful structural information from the PPS map to guide the ControlNet.
\item We demonstrate improved realism and depth preservation compared to state-of-the-art GAN-based method MI-CycleGAN~\cite{wang2024structure}, as evidenced by 40\% improvement in FID metric for C3VD (phantom) $\rightarrow$ Colon10K (clinical) translation and 20\% improvement in downstream depth prediction accuracy on the C3VD dataset for SimCol3D (simulation) $\rightarrow$ C3VD (phantom) translation.
\end{itemize}


\section{Related Work}

Colonoscopy datasets  are derived from clinical procedures~\cite{azagra2023endomapper,ma2021Colon10K}, phantom-based data~\cite{c3vd}, and fully synthetic sources~\cite{c3vd,zhang2020template}.
While synthetic and phantom datasets provide depth and pose annotations, they lack real-world complexities such as dynamic lighting, specular reflections, and tissue deformations.
Unpaired image translation using GAN-based approaches~\cite{goodfellow2020generative} with depth map~\cite{jeong2021depth,zhu2017unpaired} or mutual information~\cite{wang2024structure} has been explored to preserve structure integrity. However, GANs can be unstable, prone to model collapse, and may exhibit limited diversity. 
Recently, due to its improved mode coverage and stability~\cite{rombach2022high}, Stable Diffusion has been applied for clinical colonoscopy synthesis~\cite{du2023arsdm,xie2024ccis} and sim-to-real translation in surgical data~\cite{kaleta2024minimal,venkatesh2024surgical}, but preserving structural integrity remains a challenge~\cite{cheng2024zest,zhang2024atlantis,choi2024scribblelight}, which we address in this paper by designing a novel PPS-Controlled Diffusion model.

\section{Methods}

\begin{figure*}[tb]
    \centering
    \includegraphics[width=\textwidth]{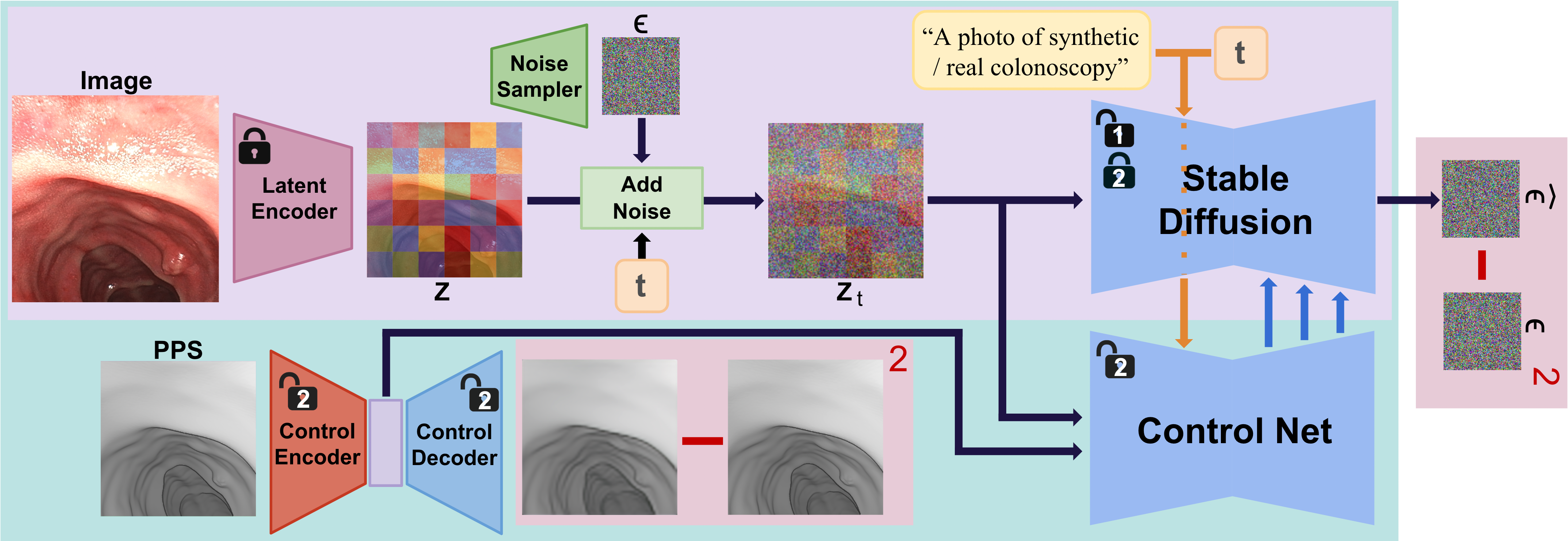}
    \caption{\small{Our method takes a depth map from any synthetic colon datasets as input and generates textures similar to real endoscopy videos while preserving the depth in the generated image. Our proposed pipeline consists of a Stable Diffusion model that can capture image statistics of both real and synthetic domains using text prompts (trained in Stage 1), and a ControlNet that guides the diffusion model to perform depth-preserving image generation through a latent encoding of a Per-Pixel Shading (PPS) map (trained in Stage 2).  
    }}
    \label{fig:pipeline}
\end{figure*}




Our goal is to perform synthetic-to-real image translation for endoscopy images while preserving the depth of the original synthetic image using a novel adaptation of Stable Diffusion~\cite{rombach2022high} and ControlNet~\cite{zhang2023adding}, as shown in Fig. \ref{fig:pipeline}. We first finetune Stable Diffusion on both the synthetic source and real target domains using text conditions to differentiate them, described in Sec.~\ref{sec:stable_diffusion}. In Sec.~\ref{sec:controlnet}, we outline our novel adaptation of ControlNet~\cite{zhang2023adding} conditioned on latent Per-Pixel Shading (PPS) features obtained using an auto-encoder that significantly improves depth-preserving image translation quality over existing techniques.



\subsection{Domain Separation using Text in Stable Diffusion}\label{sec:stable_diffusion}


When Stable Diffusion (SD) is naively finetuned on images across multiple domains, we observe that it ends up blending features between two different domains, reducing the effectiveness of adaptation. Therefore, we refine SD on both source and target domains with text-based domain separation during training and inference. To achieve this, domain separation is controlled via text conditioning: $p_{\text{source}}$ = "A photo of synthetic colonoscopy"
and
$p_{\text{target}}$ = "A photo of real colonoscopy", where $p_{source}$ and $p_{target}$ are text prompts corresponding to synthetic and real colonoscopy images respectively.
By enforcing domain separation using textual prompts, we ensure that the SD generative model maintains distinct image characteristics for each domain. Finally, we train the SD model with parameters $\theta^S$ with text prompt $p$ and image latent code $z_{t}$ at time-step $t$ using the standard loss function for denoising diffusion model:
\begin{equation}
  \mathcal{L} = \mathbb{E}_{z_{t}, \epsilon\sim\mathcal{N}(0,1),t,p}	\left[  \left\| \epsilon - \epsilon_{\theta^S}(z_{t}, t, p) \right\|^2_2\right]\,.
  \label{eq:loss_stablediffusion}
\end{equation}







\subsection{PPS-Ctrl for Depth Preservation}\label{sec:controlnet}

\begin{figure}[t]
    \centering
    \includegraphics[width=\textwidth]{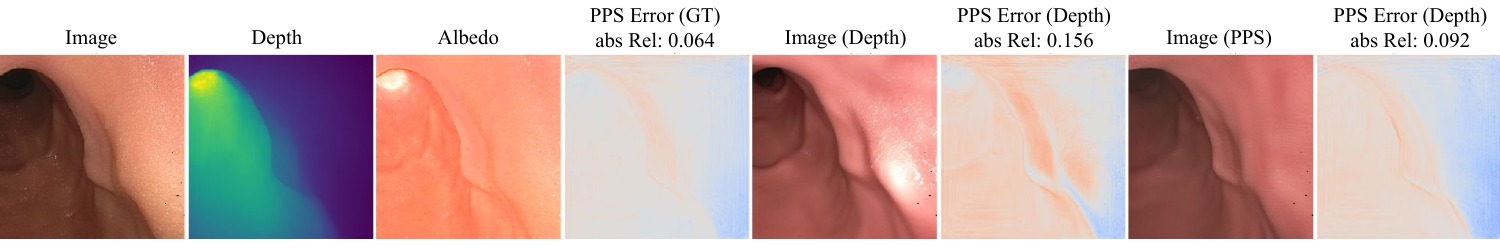}
    \caption{We calculate error between ground-truth per-pixel shading map and intensity of the original image (Col 4), image generated by depth conditioning (Col 6) and PPS condition (Col 8) to show that PPS conditioning significantly decreases structural inconsistency and closely matches that of the original image.}
    \label{fig:pps}
\end{figure}

Existing techniques for content preserving image generation~\cite{cheng2024zest,zhang2024atlantis}, often use ControlNet~\cite{zhang2023adding} conditioned on depth maps to guide the Stable Diffusion to generate depth-preserving high-quality images. However, in endoscopy, these techniques produce inconsistent surface colors for regions that are very close to (near-field) or far from (far-field) the camera and light, leading to images with a reduced perception of depth. We notice that the computed per-pixel shading field, the inner product between surface normal and incoming light at each pixel, significantly differs from the generated image intensity in near-field and far-field regions, as shown in Fig. \ref{fig:pps}. To address this issue, we propose Per-Pixel Shading (PPS)-based conditioning, which effectively captures fine-scale lighting variations and provides more stable, geometrically consistent conditioning.

Per-pixel Shading (PPS) has been utilized in Near-Field Lighting Photometric Stereo 3D reconstruction~\cite{lichy2022fast}, where the light source is positioned close to the surface, and has also proven effective for 3D estimation in endoscopy~\cite{ppsnet,beltran2024nfl,rodriguez2023lightdepth}. Building on the success of PPS in lighting-based 3D reconstruction, we propose using the PPS representation as a conditioning input for ControlNet to guide the Stable Diffusion model in preserving the source image's depth while generating realistic textures. Formally, given a synthetic depth map $D(u, v)$ and camera intrinsic $K$, we estimate the surface normal as $N(\mathbf{x}) = \frac{\partial \mathbf{x} / \partial u \times \partial \mathbf{x} / \partial v}{|\partial \mathbf{x} / \partial u \times \partial \mathbf{x} / \partial v|}$, where $\mathbf{x} = {K}^{-1} [u, v, D(u,v), 1]^\top$ is the 3D surface point corresponding to the pixel $(u, v)$ in the image plane. For a camera and a light-source located at $p_c$, each point $\mathbf{x}$ receives light with direction $L^d(\mathbf{x}) = \frac{\mathbf{x} - \mathbf{p}_c}{\|\mathbf{x} - \mathbf{p}_c\|_2}$ and intensity attenuated by inverse-square fall-off $L^a(\mathbf{x}) = \frac{1}{\|\mathbf{x} - \mathbf{p}_c\|^2_2}$, assuming no angular attenuation, similar to~\cite{lichy2022fast,ppsnet,beltran2024nfl}. Finally, we can represent Per-Pixel Shading at each surface point as:
\begin{equation}
\mathcal{PPS}(\mathbf{x}) = L^a(\mathbf{x}) \times (L^d(\mathbf{x}) ^{T}\mathbf{N}(\mathbf{x}))\,.
\end{equation}

To integrate $\mathcal{PPS}$ into the ControlNet, we use a learnable control encoder $E^C(\cdot)$, which encodes $\mathcal{PPS}$ into a feature map ($f = E^C(\mathcal{PPS})$). To further ensure the preservation of PPS-based conditioning, we introduce an additional control decoder $D^C(\cdot)$, which reconstructs the conditioning $\mathcal{PPS}$ from the feature map $f$. 

\begin{equation}
    \mathcal{L_D} = \left[ \left\| D^C(E^C(\mathcal{PPS})) - \mathcal{PPS} \right\|_2^2 \right]\,.
\end{equation}

The Control Encoder-Decoder architecture prevents feature degradation and extracts the most meaningful geometric features, preserving the control signals through later ControlNet denoising stages. We validate the importance of Control Encoder-Decoder for downstream depth map prediction tasks by training on translated data in Table \ref{tab:depth_results}. Finally, we jointly train ControlNet ($\theta^{C}$) with the PPS feature map $f$, image latent code $z_{t}$ at time-step $t$, and the text prompt $p$ as inputs, along with the Control Encoder-Decoder using the following loss:
\begin{equation}
\begin{split}
  \mathcal{L} = \mathcal{L}_{D} +
  \mathbb{E}_{z_{t}, f, \epsilon\sim\mathcal{N}(0,1),t,p}	\left[  \left\| \epsilon - \epsilon_{\theta^{C}}(z_{t}, f, t, p) \right\|^2_2\right]\,.
\end{split}
  \label{eq:loss_controlnet}
\end{equation}

To summarize, we perform Sim-to-Real translation using the following steps:\\
$\bullet$ \textit{Training, Stage 1:} Fine-tune SD on both \textit{Sim} and \textit{Real} domains using domain-specific text-prompts $p_{source}$ and $p_{target}$ optimizing eq. 1.\\
$\bullet$ \textit{Training, Stage 2:} Train ControlNet and encoder-decoder by optimizing eq. 4 on image and PPS (or depth) from \textit{Sim} domain, using SD with text-prompts corresponding to \textit{Sim} domain $p_{source}$.\\
$\bullet$ \textit{Testing:} Sample a depth map from the \textit{Sim} domain, compute the PPS using the sampled depth, and then run ControlNet + SD with text prompts corresponding to the \textit{Real} domain $p_{target}$ to generate an image that resembles the \textit{Real} domain while preserving the depth of the input \textit{Sim} domain image.



\subsection{Implementation Details}

Our Control Decoder $D^C$ consists of 4 residual blocks, and is structured with a transposed architecture of the control encoder $E^C$. We fine-tune the Stable Diffusion v2 model with a learning rate of 2e-6 and train the ControlNet for 10000 steps with a learning rate of 1e-4. We use a batch size of 16 and the AdamW optimizer for both models. For inference, we apply the DDIM scheduler and sample 20 steps.
All inputs are resized to 512 $\times$ 512.
All experiments are trained using 4xA6000 GPUs.

\begin{table}
\caption{We evaluate depth estimation accuracy on C3VD dataset by training on SimCol3D $\rightarrow$ C3VD translation data using various techniques. We consider a lower-bound performance of no translation and training on SimCol3D and upper bound performance of training on C3VD itself. We also perform an additional ablation study to prove the effectiveness of Per-Pixel Shading (PPS) over depth map for conditioning and of Control Encoder-Decoder architecture to extract more meaningful latent code.}
\label{tab:depth_results}
\centering
\renewcommand{\arraystretch}{1.2}
\setlength{\tabcolsep}{3pt}
\begin{tabular}{|l|l|ccc|}
    \hline
    \multirow{2}{*}{\begin{tabular}{@{}c@{}}Training\\Data\end{tabular}} & \multirow{2}{*}{\textbf{Condition}} & \multicolumn{3}{c|}{\textbf{DepthAnything}} \\
    \cline{3-5}
    & & RMSE $\downarrow$ & $\text{Abs}_{rel} \downarrow$ & $\delta < 1.1 \uparrow$ \\
    \hline
    SimCol3D & \textbf{Lower Bound} & 4.753 &  0.117 & 0.592 \\
    SimCol3D $\rightarrow$ C3VD & MI-CycleGAN \cite{wang2024structure} & 4.662 & 0.100 & 0.619 \\
    SimCol3D $\rightarrow$ C3VD & Ours-Depth & 4.444 & 0.095 & 0.664 \\
    SimCol3D $\rightarrow$ C3VD & Ours-Depth w. $D^C$ & 4.099 & 0.093 & 0.672 \\
    SimCol3D $\rightarrow$ C3VD & Ours-PPS & 4.135 & 0.095 & 0.663 \\
    SimCol3D $\rightarrow$ C3VD & Ours-PPS w. $D^C$ & \textbf{3.740}  & \textbf{0.088} & \textbf{0.692} \\
    C3VD & \textbf{Upper Bound} & 2.277 & 0.055 & 0.839 \\
    \hline
\end{tabular}
\end{table}

\section{Results}

\subsection{Experimental Setup}

\noindent We consider two different set of experiments. (i) SimCol3D $\rightarrow$ C3VD: to numerically evaluate depth map prediction accuracy and image translation quality on C3VD; (ii) C3VD $\rightarrow$ Colon10K: to qualitatively evaluate the image translation and depth prediction quality on real data. For ControlNet training, we reuse the train split of the source dataset.

\noindent \textbf{Datasets} We use a simulated colon dataset SimCol3D~\cite{rau2024simcol3d}, a phantom colon dataset C3VD~\cite{c3vd} , and a clinical dataset Colon10K \cite{ma2021Colon10K},  to perform source-to-targe image translation followed by depth map prediction on the target domain. 

\noindent \textbf{Metrics} We evaluate the image translation quality using Freceht Inception Distance (FID) between the source-to-target translated data and the original target data. We evaluate the depth estimation performance using root-mean-squared error (RMSE), absolute relative error (AbsRel), and percentage of pixels within 10\% of the actual depth values $\delta$< 1.1.

\noindent \textbf{Baselines} We compare with a state-of-the-art depth-preserving image-to-image translation algorithm for endoscopy, MI-CycleGAN~\cite{wang2024structure}.

\subsection{Depth Estimation}

To assess the structure-preserving capability of our image translation, we compare the performance of multiple fine-tuned DepthAnything~\cite{yang2024depth} models. More specifically, we train each model using translated images paired with the corresponding depth maps from the original source data then test on images from the target domain. In Table \ref{tab:depth_results}, we compare the depth estimation accuracy of training on our translated data against translation data using the existing approach, MI-CycleGAN~\cite{wang2024structure}, for SimCol3D $\rightarrow$ C3VD. 
We also evaluate a model trained on only non-translated SimCol3D data and evaluated on C3VD data as lower bound and a model trained and tested on C3VD data as upper bound. 
We observe that our approach significantly outperforms MI-CycleGAN~\cite{wang2024structure} with 20\% improvements in RMSE and similar improvements in $Abs_{rel}$ and $\delta< 1.1$. 

\begin{figure*}[tb]
    \centering
    \includegraphics[width=\textwidth]{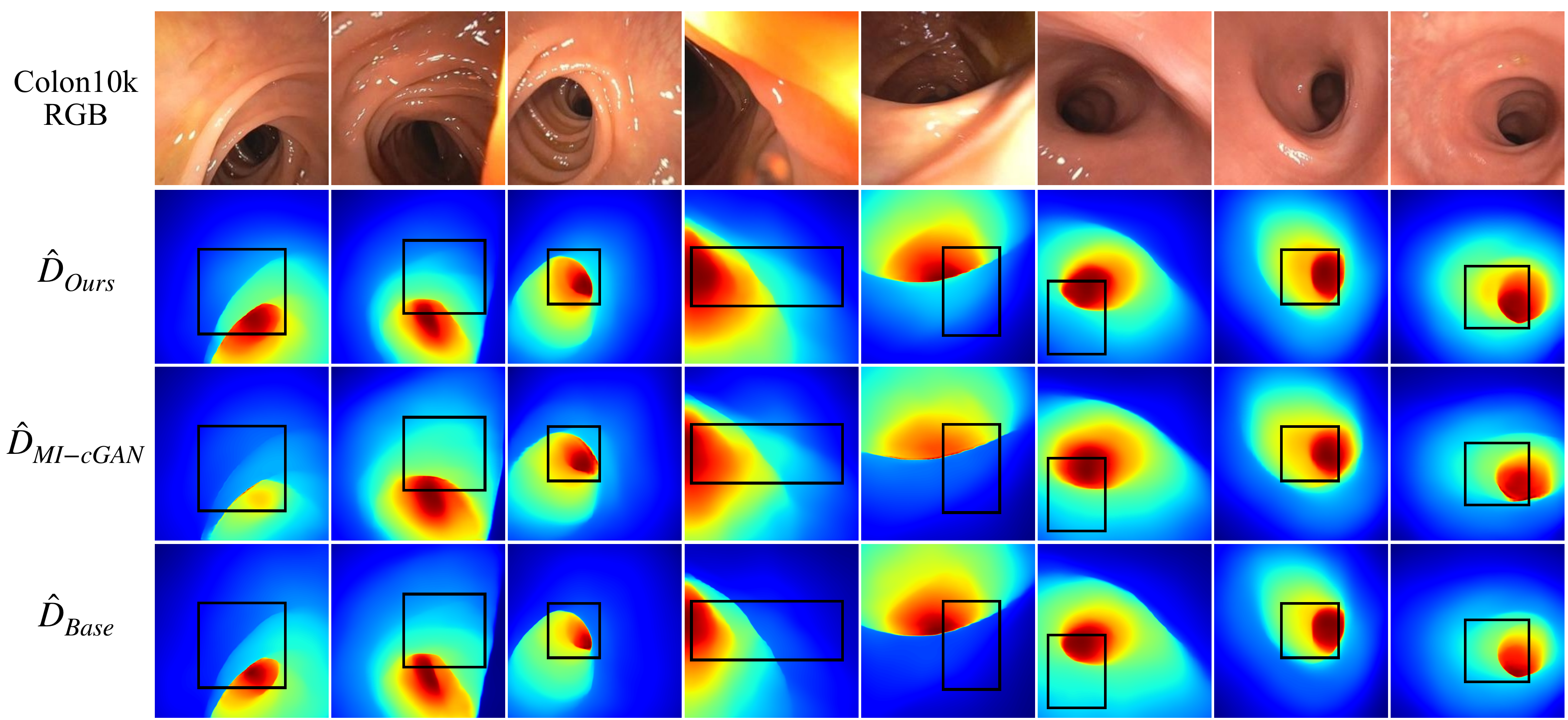}
    \caption{Comparison of depth estimation on the Colon10K dataset using DepthAnything~\cite{yang2024depth}, trained on C3VD$\rightarrow$Colon10K translations from our method ($\hat{D}_{Ours}$) and MI-CycleGAN \cite{wang2024structure} ($\hat{D}_{MI-cGAN}$), as well as models trained on C3VD alone without translation ($\hat{D}_{Base}$). Our translated data significantly enhances depth estimation compared to MI-CycleGAN and no translation, particularly in the regions highlighted by the black box.}
    \label{fig:depth estimation gt}
\end{figure*}
\vspace{-0.5em}
 
In the absence of ground-truth depth annotations for clinical data, we qualitatively assess the impact of translating C3VD $\rightarrow$ Colon10K on depth prediction (Fig.\ref{fig:depth estimation gt}).
When trained on our method's translated data over non-translated data, the resulting depth maps capture the colon’s contours more effectively in regions close to the camera (highlighted in both \textit{oblique} and \emph{en face} frames).
Furthermore, when comparing models trained on our translated images with those trained on MI-CycleGAN\cite{wang2024structure} translations, we observe similar improvements in accurately representing more distant regions of the colon (particularly evident in  \textit{down-the-barrel} views).

\begin{figure*}[tb]
    \centering
    \includegraphics[width=\textwidth]{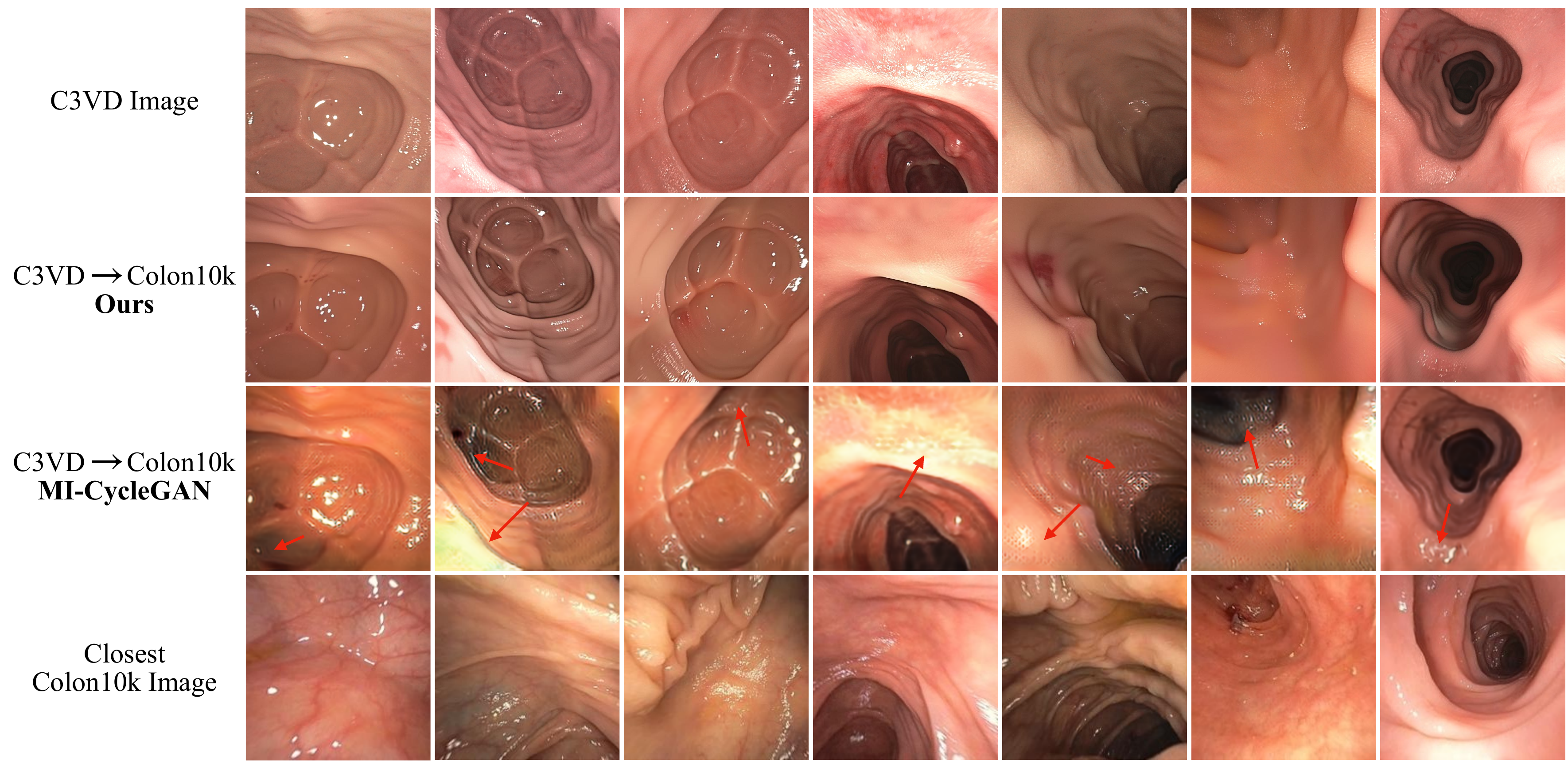}
    \caption{Comparison of C3VD $\rightarrow$ Colon10K translation results between our method and MI-CycleGAN \cite{wang2024structure}, with similar views from Colon10K provided for reference. MI-CycleGAN often fails to preserve depth, introducing incorrect dark textures that mimic down-the-barrel views (Columns 1, 2, 3, 6) and generating unrealistic reflections (Columns 2, 4, 5, 7). Additionally, it produces noticeable checkerboard artifacts (best viewed when zoomed in), which our approach effectively mitigates.}
    \label{fig:rgb}
\end{figure*}

\subsection{Image Translation}

We demonstrate the effectiveness of domain translation in Table \ref{tab:fid_kid} using image translation quality measured by FID.
We consider two different scenarios for translation (SimCol3d $\rightarrow$ C3VD and C3VD $\rightarrow$ Colon10K) for our method, MI-CycleGAN \cite{wang2024structure}, and no translation. 
Our translation enhances texture realism (Fig, \ref{fig:rgb}, has fewer artifacts such as over specularity and decoloration, and preserves source image structure more than the MI-CycleGAN \cite{wang2024structure} baseline.




\begin{table}[h]
    \centering
    \caption{We evaluate the image translation quality using FID (lower is better) of SimCol3D $\rightarrow$ C3VD and C3VD $\rightarrow$ Colon10K for 3 methods: no translation, MI-CycleGAN~\cite{wang2024structure}  translation, and Our translation. Note that this only measures texture translation quality and ignores structure preservation. 
    }
    \label{tab:fid_kid}
    
\begin{tabular}{|l|ccc|ccc|}
    \hline
    \multirow{2}{*}{\begin{tabular}{@{}c@{}}Translation \\ Algorithm\end{tabular}} & \multicolumn{3}{c|}{\textbf{SimCol3D $\rightarrow$ C3VD}} & \multicolumn{3}{c|}{\textbf{C3VD $\rightarrow$ Colon10K}} \\
    \cline{2-7}
    & No Trans. & MI-cGAN & Ours & No Trans.& MI-cGAN & Ours \\
    \hline
    \textbf{FID $\downarrow$} & 0.545 &  0.591 & \textbf{0.437} & 0.527 & 0.498 & \textbf{0.297} \\
    \hline
\end{tabular}
\end{table}
\vspace{-1.5em}

\subsection{Ablation Study}

We show the effectiveness of our key contributions: (i) using a PPS map over a depth map for ControlNet condition and (ii) using Control Encoder-Decoder architecture to extract more meaningful structure-preserving latent codes over only using a latent encoder in ControlNet. We use downstream depth estimation task on SimCol3D $\rightarrow$ C3VD and present the result in Table \ref{tab:depth_results}. The results show that replacing depth maps with PPS consistently improves RMSE, absolute relative error, and $\delta$ < 1.1 metrics, confirming PPS provides richer geometric cues for better structural preservation. Integrating a decoder $D^C$ into ControlNet further enhances depth estimation performance across all conditions. Combining these two advances, we obtain the best performance with Ous-PPS with $D^C$ setting.


\vspace{-0.5em}
\subsection{Limitations}
While our method learns to add specularity caused by the mucus layer during Sim-to-Real transfer, we can not control the amount of specular reflections as we do not explicitly model it using physically-based material modeling. Our model also struggles to preserve or add fine features, such as small blood vessels (see Fig.\ref{fig:rgb}, row 4, column 1), which can be addressed in the future with explicit conditioning.

\section{Conclusions}

We present PPS-Ctrl, a sim-to-real image translation framework using Stable Diffusion with ControlNet, conditioned on a Per-Pixel Shading (PPS) to enhance structural preservation and realism in colonoscopy images. 
Our experiments demonstrate that PPS-Ctrl outperforms existing texture transfer methods in achieving photorealistic results. Moreover, by training depth estimation models on our translated images, we observed improved performance on out-of-domain clinical data, underscoring the effectiveness of our approach in bridging the domain gap between synthetic and real endoscopic datasets.

\section{Acknowledgement}

This work is supported by a National Institute of Health (NIH) project 
\#1R21EB035832 “Next-gen 3D Modeling of Endoscopy Videos”.

\renewcommand{\normalsize}{\fontsize{10pt}{12pt}\selectfont} 
\bibliographystyle{splncs04}
\bibliography{references}

%






\end{document}